# A Deep Convolutional Neural Network for Lung Cancer Diagnostic

Mehdi Fatan Serj*, Bahram Lavi†, Gabriela Hoff‡
and Domenec Puig Valls§

## Abstract

In this paper, we examine the strength of deep learning technique for diagnosing lung cancer on medical image analysis problem. Convolutional neural networks (CNNs) models become popular among the pattern recognition and computer vision research area because of their promising outcome on generating high-level image representations. We propose a new deep learning architecture for learning high-level image representation to achieve high classification accuracy with low variance in medical image binary classification tasks. We aim to learn discriminant compact features at beginning of our deep convolutional neural network. We evaluate our model on Kaggle Data Science Bowl 2017 (KDSB17) data set, and compare it with some related works proposed in the Kaggle competition.

## 1 Introduction

According to the World Health Organization (WHO) the lung cancer is classified as a noncommunicable disease and it is the $5^{\text{th}}$ cause of death (associated to trachea and bronchus cancer) among of all possible causes evaluated in the world [13]. Lung cancer is the most common type of cancer affecting the lives of many people worldwide for several decades, which causes of death estimated for nearly 1.59 million people per year. According to Siegel et al. [15] is expected for 2017 that lung cancer, among all cancers, presents the highest incidence for new cases (222,500 cases) considering men and women as one group, and the highest mortality (155,870 deaths) for both genders in United States. Among men and women it will be the second cause of death, losing just for prostate cancer for men and for breast cancer for women [15]. Considering all possible causes of death it is observed, that lung cancer have higher incidence on North America and Europe [4,13]. It is reported that 58% of the lung cancers occurred in less developed countries [4,6]. According to [6] lung cancer cause one in five cancer deaths worldwide (1.6 million deaths, 19.4% of all cancer deaths). Considering this statistics and the possibility of increase the life quality or cure for early detection [12], it is important to invest on systems for early lung cancer detection. The National Lung Screening Trial have demonstrated reduction in mortality of 20% considering a low-dose computed tomography screening [12]. Usually when lung cancer became symptomatic it is in advance stage [12,19], so to invest on risk groups screening could reduce the number of deaths [19]. One can increase the diagnostic performance and the survival of patients [9] by associating a computed-aided diagnostic (CAD) system to the standard diagnostic procedure [12,19]. Different approaches to CAD systems have been applied using neural networks [3,9,17] showing promising results. In this manner, medical image analysis plays an important role of early detection such kind of cancer from medical images. This field of research can be typically categorized in two kinds of task: segmentation; to determine a suspicious region which identified as cancer, and classification; that uses to diagnosis and/or assessment of a kind of cancer, and in some case, uses for whether a person is under the cancer risk or not (*aka binary classification*).

---

*Dept. of Computer Engineering and Maths, Universitat Rovira i Virgili. Email: mehdi.fatanserj@estudiants.urv.cat
†Dept. of Electrical and Electronic Engineering, University of Cagliari. Email: lavi.bahram@diee.unica.it
‡Dept. of Physics, University of Cagliari. Email: Gabriela.Hoff@ca.infn.it
§Dept. of Computer Engineering and Maths, Universitat Rovira i Virgili. Email: domenec.puig@urv.cat

There has been an increasing interest among the researchers in medical image to develop lung cancer diagnosis methods based on deep learning techniques. Deep learning methodology is an improved version of the artificial neural networks, which consists of several layers to generate high-order features from its input, and then, brings out the predicted value on the top of the network. Among deep learning techniques, in particular, convolutional neural networks (CNNs) have been widely applied in computer vision tasks. The results of CNNs have proven its robustness in object recognition localization in variant images. Recently, the applied deep learning techniques are emerged also in medical image analysis task. The bulk of interesting deep learning works, proposed to improve the performance of medical images, have been applied either by modifying the architecture of the existing deep learning networks, or proposing new ones.

Motivated by the success of deep learning, our work targets to learn a high-level and discriminative feature using by CNN to diagnose of a lung cancer. To address this issue, we propose a new deep convolutional neural network (dCNN) architecture. Our network comprised by three convolution layers, two max-pooling layers, a full-connected layer, and a binary soft-max layer. This network begins with two sequential convolution layers in order to generate high-order feature at the preliminary layers. Then the generated feature maps pass through the other layers to the top of the network. To validate the performance of our dCNN, we test it on KDSB17 data set. The experimental results show that our dCNN perform impressively, in fact, it significantly outperforms many all the Kaggle competitor's method in lung cancer diagnosis image analysis task.

This paper is structured as follows. We first summarize the related work in Sect. 2 In Sect. 3 we discuss our proposed deep learning architecture for lung cancer classification. In Sect. 4 we evaluate our model effectiveness on the KDSB17 data set, and compare it with some first ranked positions on Kaggle competition.

## 2 State-of-the-art

In this section we describe some of the existing lung cancer diagnostic approaches to classification problems. In [2], CADe and CADx systems based on Houns eld unit and perform the segmentation by combining region growing algorithm and morphological filters. A powerful learning model proposed in [3], which employed back-propagation neural network for classification that would classify the digital X-ray, CT-images, MRIs, etc as cancerous or non-cancerous. Further, A genetic algorithm has been used that extracts feature on the basis of the fitness function. The work presented in [8], briefly examine the potential use of classification based data mining techniques such as *Rule* based, Decision tree, Naive Bayes and Artificial Neural Network to massive volume of health-care data. The accuracy on the prediction comparing the two methods is presented. A multi-stage framework proposed in [10] detects nodules in 3D lung CAT scans, determines if each nodule is malignant, and assigns a cancer probability based on these results. In [14], a Histogram Equalization is used for pre-processing of the images and feature extraction process and neural network classifier to check the condition of a patient in its early stage whether it is normal or abnormal.

The work by Song et al. [16] presents a cancer classification method by investigating on three types of deep neural networks (e.g., CNN, DNN, and SAE) which have been designed particularly for lung cancer classification. Those networks are applied to the CT image classification task with some modifications for the benign and malignant lung nodules. Another work for cancer classification in [18] which developed an automated classification scheme for lung cancer presented in microscopic images using a deep convolutional neural network (dCNN), which is a major deep learning technique. The dCNN used for classification consists of three convolutional layers, three pooling layers, and two fully connected layers. Moreover, Kuruvilla et al. [11] proposed a model forcancer classification. The statistical parameters like mean, standard deviation, skewness, kurtosis, fifth central moment and sixth central moment are used for classification. The classification process is done by feed forward and feed forward back propagation neural networks. Compared to feed forward networks the feed forward back propagation network gives better classification.

Taher et al. [17] presents two segmentation methods, Hopfield Neural Network (HNN) and a Fuzzy C-Mean (FCM) clustering algorithm, for segmenting sputum color images to detect the lung cancer in its early stages. The manual analysis of the sputum samples is time consuming, inaccurate and requires intensive trained person to avoid diagnostic errors. A Computer-aided diagnosis (CAD) system proposed in [9] which uses deep features extracted from an auto-encoder to classify lung nodules as either malignant or benign. This work got the $3^{\text{rd}}$ ranked position of the Kaggle lung cancer competition [7]. To the best of our knowledge, it is also the only paper which presents the results with appropriate comparison of the ones who have been competed in Kaggle competition. We therefore present our experimental evaluation by following their experimental strategy.

## 3 Proposed Network

In this section, we present the proposed dCNN architecture in details. We first describe the overall framework of dCNN, and then explain our loss function used to train the proposed dCNN model.

## 3.1 The Overall network

The proposed dCNN architecture mainly consists of the following layers: four convolution layers which follow two max-pooling layers, a full-body convolution layer, and one fully connected layer with two softmax units. As shown in Fig. 1 the network begins with two convolution layers, in which the first convolution layer takes the image with input size of $120 \times 120$ pixels. The first convolution layer consists of 50 feature maps with the convolution kernel of $11 \times 11$. The second convolution layer consists of 120 feature maps with the convolution kernel of $5 \times 5$. And the last convolution layer consists of 120 feature maps with the convolution kernel of $3 \times 3$. The kernel size for max pooling layers is $2 \times 2$ and the stride of 2 pixels, and the fully-connected layer generates an output of 10 dimensions. These 10 outputs are then passed to another fully connected layer containing 2 softmax units, which represent the probability that the image is containing of the lung cancer or not. Note that each convolution layer in our dCNN model is followed by a rectified linear unit (ReLU) layer to produce their outputs.

## 3.2 Loss function

We use *cross-entropy* as the loss function of our training model which is computed by using multinomial logistic regression objective. This aims to maximize the probability of the patinets with cancer by maximizing the multinomial logistic regression objective. At this paper, we achieve this by minimizing the cross-entropy loss for each training sample. The training aims to maximize the probability of the correct person identity by maximizing the multinomial logistic regression objective. Therefore, this is known as softmax loss function in our model. For a single input sample $x$ and a single output node $i$ in the last layer, the loss could be calculated by

$$p(y = i|x) = \frac{e^{a_i(x)}}{\sum_{i=1}^{K} e^{a_i(x)}} \quad (1)$$

where $p(x)$ is approximated maximum-function, $a(x)$ is the activation function of $x \in M$ with $M \subset \mathbb{Z}^2$, and $K$ is the number of classes (which is equal 2 at this work). And finally, the gradients are computed by standard backpropagation of the error [5].

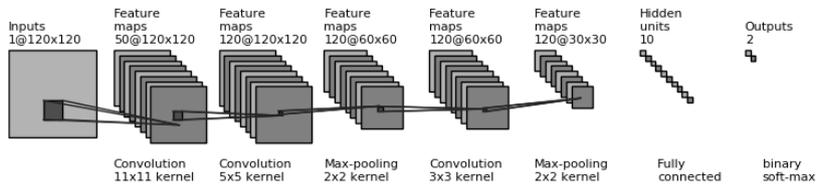

Figure 1: Scheme of our proposed dCNN architecture.

## 3.3 Performance measurements

The performance of a method, in medical image analysis, is typically measured by specificity, sensitivity, and F1 score.

Sensitivity measures the proportion of actual positives samples that correctly identify, in which the percentage of cancerous nodule that is *correctly* classified as cancerous. Therefore, it is computed according to the following definition:

$$sensitivity = \frac{TP}{TP + FN} \qquad (2)$$

where $TP$ (*true positive*) is the number of nodules which have been successfully detected, and $FN$ (*false negative*) is the number of nodules which have been detected by the method. In contrast, specificity measures the proportion of identified negatives samples, in which the percentage without cancerous nodule is correctly classified as non cancerous. In this manner, specificity is computed as:

$$specificity = \frac{TN}{TN + FP} \qquad (3)$$

where $TN$ (*true negative*) is the number of non-cancer patients which have been successfully classified, and $FP$ (*false positive*) is the number of non-cancer patients which have been wrongly classified as cancer.

And, F1-score measures the average F1 score through different class labels which is computed as:

$$F1 = 2 \times \frac{PPV \times TPR}{PPV + TPR} \qquad (4)$$

where PPV

$$PPV = \frac{TP}{TP + FP} \qquad (5)$$

and TPR

$$TPR = \frac{TP}{TP + FN} \qquad (6)$$

As presented at the Kaggle competition, all results by competitors should be given as the performance measurment of *log-loss* function as follow:

$$loss(p, q) = -\frac{1}{|M|} \sum_{c \in M} w(c).p(c)logq(c) \qquad (7)$$

where p is the true distribution, $f_c$ is the frequency of class $c$ in the mini-batch, and

$$w(c) = \begin{cases} \frac{f_{cancer-free}}{f_{cancer}} & \text{if c belongs to the cancer class} \\ 1 & \text{otherwise} \end{cases}$$

## 4 Experimental evaluation

We evaluated our dCNN model on Kaggle Data Science Bowl 2017 (KDSB17) data set, which recently have been released for a competition on lung cancer images. We also compared our result with different competitors by presenting their results.

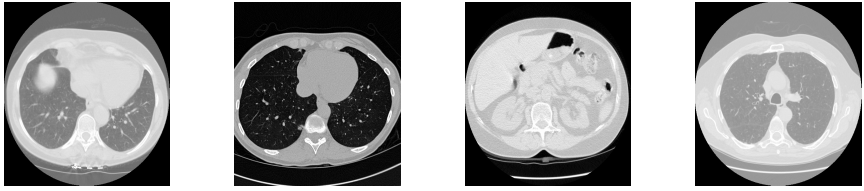

Figure 2: Four slices of different patient chests cavity from the Kaggle Data Science Bowl 2017 data set.

## 4.1 Data set for training and testing

We evaluate our dCNN model on the data set provided in Kaggle Data Science Bowl 2017 (KDSB17) which contains CT scan images of patients, as well as their cancer status. All the images of this data set are in size of $512 \times 512$ pixels. This dataset includes CT scans of lung cancer patients and also non patients, which subdivided into the two parts: (i) part 1 generated to be used for nodule detection and localization, while part 2 used for recognition of CT images as patient or not. In this paper we used part 2 for implementation of our dCNN for lung cancer diagnostic. This part of the data set consists of 63890 lung images of patients with cancer, and 171345 images of non cancers. Fig. 4.1 contains sample slices from this data set.

We used 50% of the images for training , 25% used for cross-validate set, and remains used to test our model. We have implemented dCNN by using Matlab 2017b with Deep Learning toolbox by running on an ASUS n552vw laptop which has NVIDIA GeForce GTX 960M GPU. Note that we re-scaled all the images of this data set to the size of $120 \times 120$ pixels before our training stage, because of memory space limitation on GPU to train our dCNN.

## 4.2 Implementation details

Training data images are randomly divided into mini-batches. The model performs forward propagation on the each mini-batch and computes the output and loss. Then, backpropagation is used to compute the gradients on this batch, and network weights are updated. We perform stochastic gradient descent [1] to perform weight updates, and the loss functions (see Sect. 3.2) is minimized by stochastic gradient descent with the size of batch 128 which has been done on 11000 iterations on the sample. Also, we experiment a few hyperparameters and choose the learning rate 0.001, and use a momentum of $\mu = 0.9$.

## 4.3 Results and discussion

In medical image analysis, the performances of a method is typically measured by using sensitivity (the true positive rate) and specificity (the true negative rate) instead of measuring direct accuracy of the classification. In the KDSB17, the competitors were evaluated their results using the log-loss metric. By following the experimental results presented in [9], and to assess overall classification relevance, we therefore compute the F1-score, too.

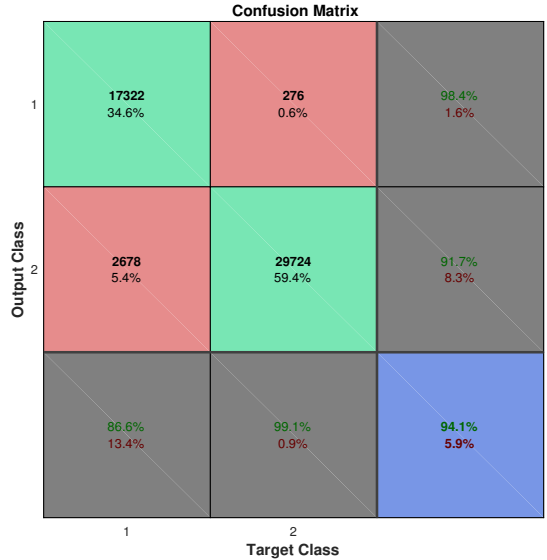

Figure 3: Confusion matrix of proposed method in lung cancer diagnostic task.

Table 1 reports the overall results with three mentioned metrics (see Sect. 3.3). Note that we only compare the results of sensitivity, specificity, and F1-score with AIDA team [9] because the results from the others have not been reported by using theses performance measurements, but only log-loss result. However, table 2 gives the comparison the log-loss results of our proposed model with other competitors of Kaggle competition.

Table 1: The result of our proposed dCNN in three metric measurements.

|          | *Sensitivity* | *Specificity* | *F1* |
|----------|---------------|---------------|------|
| **AIDA [9]** | 0.538 | 0.648 | 0.33 |
| **ours** | **0.87** | **0.991** | **0.95** |

In the KDSB17, all methods compared by log-loss, so in table 2, we present three most advance methods in literature sorted by this measurement. As can be seen in the following table, our method is the best in comparison with the winner methods of this Kaggle competition. Our log-loss is 0.20 in compare with grt123 which is 0.39 without consideration of three other metrics. Sensitivity and specificity are very significant considering to table 1, but the reason of very high specificity can be higher number of Cancer-free CT images relative to cancer ones. For more detail evaluation of our proposed dCNN model, we also present the confusion matrix in Fig. 4.2 which reports efficiency of our model in different aspects of recognition rate and error estimation for each class including extra scores such as precision and sensitivity.

Fig. 4.3 shows accuracy and loss of dCNN during training phase witch illustrates how loss error decreases during the optimizing network parameters and how accuracy increases and oscillation is because of using mini-batch instead of full-batch in deep neural networks such as dCNN. This epoch number or iterations have been chosen utilizing Fig. 4.3 which shows in what steps validation loss is minimum and also overfitting is happened and helps to specify appropriate value. This suitable epoch number which is a good evidence of high generalization ability of dCNN and prevents of movement from generalization to memorizing. Moreover, this fact leads to achieve a trade-off between overfitting and underfitting. The high number of CT data allowed us to employ a moderate mini-batch size and regulate training process to gain appropriate speed and efficiency.

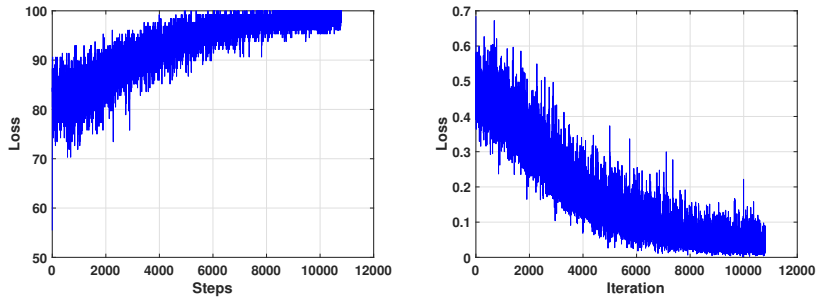

Figure 4: (left): the recognition rate of training set, and (right): total loss function as gradient steps which are taken with respect to mini-batches.

## 5 Conclusion

In this paper, we proposed a deep convolutional neural network architecture for binary classification of lung cancer that had aimed to learn high-order convolutional features at the initial layers of the model. The results presented in this work have been compared with the obtained by some of the high-ranked methods in Kaggle competition, and our model has got the lowest logloss value with respect to the other teams. The proposed dCNN architecture not only proved the power of itself, but it also proved that the features in the CT-scan images can be learned and compacted at the preliminary layers of a deep model. In terms of future work, we plan to extend the proposed deep models capabilities by integrating the automatic segmentation of nodules in the above-mentioned data set.

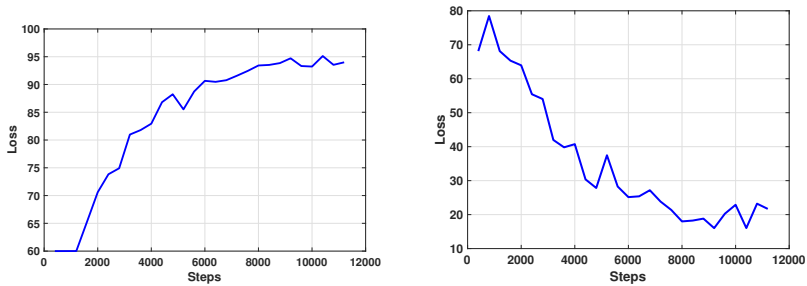

Figure 5: (left): the recognition rate of cross-validation set, and (right): total loss function as gradient steps which are taken with respect to mini-batches.

Table 2: Comparison of our result with three winner teams on KDSB17 data set.

| Rank | Name | Log-Loss |
|---|---|---|
| 1 | grt123 | 0.3997 |
| 2 | JuliandeWit&DanielHammack | 0.4011 |
| 3 | Aidence | 0.4012 |
| 41 | AIDA | 0.5271 |
| 50 | Excelsior | 0.5504 |
| 100 | rnrq | 0.6049 |
| 200 | Byeong-wooJeon | 0.6195 |
|  | **Ours** | **0.2098** |

# References

[1] L. Bottou. Stochastic gradient tricks. *Neural Networks, Tricks of the Trade, Reloaded*, 7700:430–445, 2012.

[2] K. Doi. Computer-aided diagnosis in medical imaging: historical review, current status and future potential. *Computerized medical imaging and graphics*, 31(4):198–211, 2007.

[3] M. J. DCruz, M. A. Jadhav, M. A. Dighe, M. V. Chavan, and J. Chaudhari. Detection of lung cancer using backpropagation neural networks and genetic algorithm.

[4] J. Ferlay, I. Soerjomataram, M. Ervik, R. Dikshit, S. Eser, C. Mathers, M. Rebelo, D. Parkin, D. Forman, and F. Bray. Cancer incidence and mortality worldwide: Iarc cancerbase no. 11 [internet]. lyon, france: International agency for research on cancer. globocan. 2013; 2012 v1. 0. *Available from: Available from: http://globocan. iarc. fr*, 2015.

[5] R. Hecht-Nielsen et al. Theory of the backpropagation neural network. *Neural Networks*, 1(Supplement-1):445–448, 1988.

[6] A. Jemal, P. Vineis, F. Bray, L. Torre, and D. Forman. *The cancer atlas*. American Cancer Society, 2016.

[7] Kaggle. Data science bowl 2017, 2017.

[8] V. Krishnaiah, D. G. Narsimha, and D. N. S. Chandra. Diagnosis of lung cancer prediction system using data mining classification techniques. *International Journal of Computer Science and Information Technologies*, 4(1):39–45, 2013.

[9] K. Kuan, M. Ravaut, G. Manek, H. Chen, J. Lin, B. Nazir, C. Chen, T. C. Howe, Z. Zeng, and V. Chandrasekhar. Deep learning for lung cancer detection: Tackling the kaggle data science bowl 2017 challenge. *arXiv preprint arXiv:1705.09435*, 2017.

[10] D. Kumar, A. Wong, and D. A. Clausi. Lung nodule classification using deep features in ct images. In *Computer and Robot Vision (CRV), 2015 12th Conference on*, pages 133–138. IEEE, 2015.

[11] J. Kuruvilla and K. Gunavathi. Lung cancer classification using neural networks for ct images. *Computer methods and programs in biomedicine*, 113(1):202–209, 2014.

[12] D. E. Midthun. Early detection of lung cancer. *F1000Research*, 5, 2016.

[13] W. H. Organization et al. Global health observatory (gho) data 2015. *Retrieved*, 3:23, 2015.

[14] R. S. Shriwas and A. D. Dikondawar. Lung cancer detection and prediction by using neural network. *International Journal of Electronics & Communication (IIJEC)*, 3(1), 2015.

[15] R. L. Siegel, K. D. Miller, S. A. Fedewa, D. J. Ahnen, R. G. Meester, A. Barzi, and A. Jemal. Colorectal cancer statistics, 2017. *CA: a cancer journal for clinicians*, 67(3):177–193, 2017.

[16] Q. Song, L. Zhao, X. Luo, and X. Dou. Using deep learning for classification of lung nodules on computed tomography images. *Journal of Healthcare Engineering*, 2017, 2017.

[17] F. Taher and R. Sammouda. Lung cancer detection by using artificial neural network and fuzzy clustering methods. In *GCC Conference and Exhibition (GCC), 2011 IEEE*, pages 295–298. IEEE, 2011.

[18] A. Teramoto, T. Tsukamoto, Y. Kiriyama, and H. Fujita. Automated classification of lung cancer types from cytological images using deep convolutional neural networks. *BioMed Research International*, 2017, 2017.

[19] G. Veronesi, P. Maisonneuve, L. Spaggiari, C. Rampinelli, A. Pardolesi, R. Bertolotti, N. Filippi, and M. Bellomi. Diagnostic performance of low-dose computed tomography screening for lung cancer over five years. *Journal of Thoracic Oncology*, 9(7):935–939, 2014.